\documentclass[lettersize,journal]{IEEEtran}
\usepackage[T1]{fontenc}

\usepackage{newtxtext,newtxmath}
\usepackage{amsmath,amsfonts}
\usepackage{multirow,bigstrut}
\usepackage{algorithmic}
\usepackage{array}
\usepackage[caption=false,font=normalsize,labelfont=sf,textfont=sf]{subfig}
\usepackage{textcomp}
\usepackage{stfloats}
\usepackage{url}
\usepackage{verbatim}
\usepackage{graphicx}
\usepackage{cite}
\usepackage{ragged2e}
\usepackage{booktabs,makecell,multirow,tabularx}
\usepackage{hyperref}
\usepackage{xcolor}
\usepackage[ruled,linesnumbered]{algorithm2e}
\usepackage{fancyhdr}

\begin{document}
	\fancypagestyle{firstpagefooter}{
		\fancyhf{}
		\renewcommand{\headrulewidth}{0pt}
		\renewcommand{\footrulewidth}{0pt}
		\fancyfoot[C]{\footnotesize Copyright © 2024 IEEE. Personal use of this material is permitted. However, permission to use this material for any other purposes must be obtained from the IEEE by sending an email to pubs-permissions@ieee.org.}
	}
	
	\title{Excavate the potential of Single-Scale Features: A Decomposition Network for Water-Related Optical Image Enhancement}
	\author{
		\IEEEauthorblockN{
			Zheng Cheng, Wenri Wang, Guangyong Chen, Yakun Ju,~\IEEEmembership{Member,~IEEE}, Yihua Cheng, Zhisong Liu,~\IEEEmembership{Member,~IEEE}, Yanda Meng, Jintao Song
		}
		\thanks{This work was supported in part by the National Natural Science Foundation of China under Grants 62173091 and 62073082; in part by the Natural Science Foundation of Fujian Province under Grants 2024J09021 and 2023J01268; \textit{(Corresponding author: Jintao Song.)}
		}
		\thanks{Zheng Cheng is with the College of Computer Science \& Technology, Qingdao University, Qingdao 266071, China.(2022020691@qdu.edu.cn).}
		\thanks{Wenri Wang and Guangyong Chen are with the College of Computer and Data Science, Fuzhou University, Fuzhou 350108, China.(cgykeda@mail.ustc.edu.cn, 241027144@fzu.edu.cn)}
		\thanks{Yakun Ju is with School of Computing and Mathematical Sciences, University of Leicester, Leicester, LE1 7RH, UK.(yj174@leicester.ac.uk)}
		\thanks{Yihua Chen is with School of Computer Science, University of Birmingham, Birmingham, B15 2TT, UK.(y.cheng.2@bham.ac.uk)}
		\thanks{Zhisong Liu is with Department of Computational Engineering, Lappeenranta-Lahti University of Technology (LUT), and Atmospheric Modelling Centre-Lahti. 53850 Finland.(zhisong.liu@lut.fi)}
		\thanks{Yanda Meng is with Computer Science Department, University of Exeter, Exeter, EX4 4QJ, UK.(y.m.meng@exeter.ac.uk)}
		\thanks{Song Jintao  is with the Computer Science and Technology at Shandong Technology and Business University, Yantai 264003, China(202414177@sdtbu.edu.cn)}
		
	}

	\markboth{Journal of \LaTeX\ Class Files,~Vol.~14, No.~8, August~2021}%
	{Shell \MakeLowercase{\textit{et al.}}: A Sample Article Using IEEEtran.cls for IEEE Journals}

	\IEEEpubid{}
	\maketitle
	\thispagestyle{firstpagefooter}

	\begin{abstract}
		
		Underwater image enhancement (UIE) techniques aim to improve visual quality of images captured in aquatic environments by addressing degradation issues caused by light absorption and scattering effects, including color distortion, blurring, and low contrast. Current mainstream solutions predominantly employ multi-scale feature extraction (MSFE) mechanisms to enhance reconstruction quality through multi-resolution feature fusion. However, our extensive experiments demonstrate that high-quality image reconstruction does not necessarily rely on multi-scale feature fusion. Contrary to popular belief, our experiments show that single-scale feature extraction alone can match or surpass the performance of multi-scale methods, significantly reducing complexity.
		
		To comprehensively explore single-scale feature potential in underwater enhancement, we propose an innovative Single-Scale Decomposition Network (SSD-Net). This architecture introduces an asymmetrical decomposition mechanism that disentangles input image into clean layer along with degradation layer. The former contains scene-intrinsic information and the latter encodes medium-induced interference. It uniquely combines CNN's local feature extraction capabilities with Transformer's global modeling strengths through two core modules: 1) Parallel Feature Decomposition Block (PFDB), implementing dual-branch feature space decoupling via efficient attention operations and adaptive sparse transformer; 2) Bidirectional Feature Communication Block (BFCB), enabling cross-layer residual interactions for complementary feature mining and fusion. This synergistic design preserves feature decomposition independence while establishing dynamic cross-layer information pathways, effectively enhancing degradation decoupling capacity. Compared with traditional multi-scale approaches, our single-scale design significantly reduces redundant computations and model parameters without sacrificing enhancement quality.
		
	\end{abstract}
	

	\begin{figure}[t]
		\centering
		\includegraphics[width=0.48\textwidth]{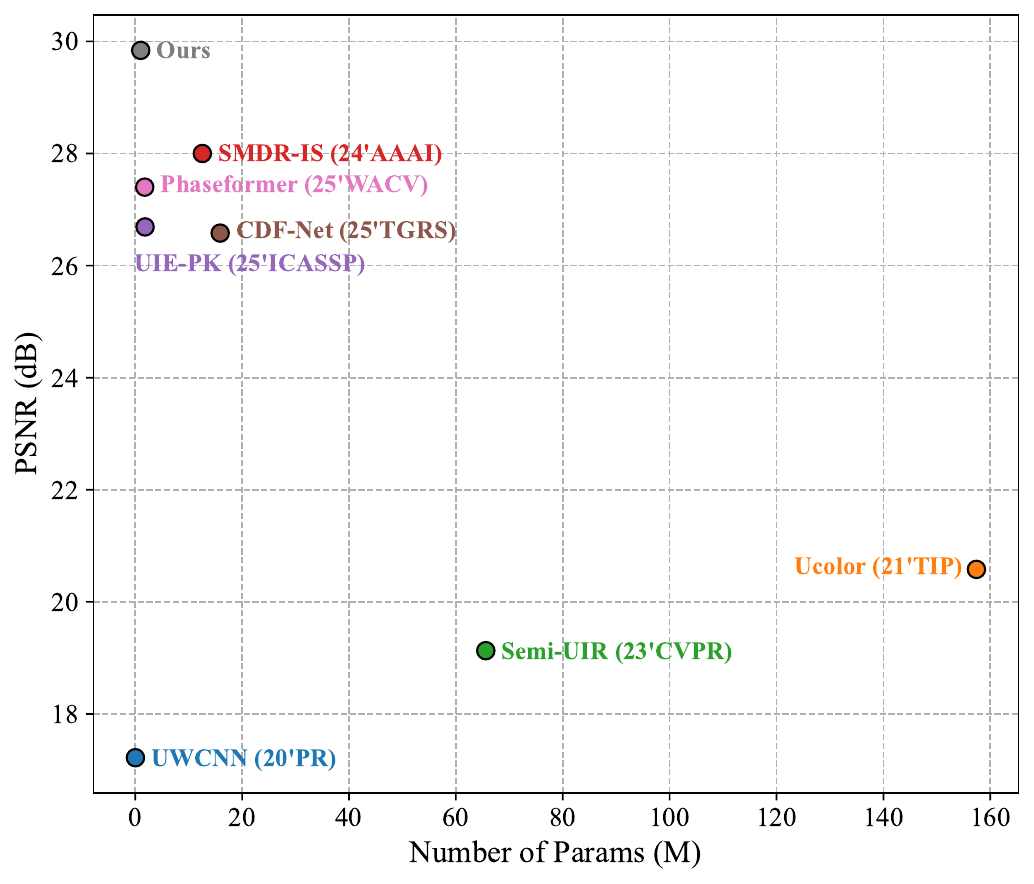}
		\caption{Comparison of the number of model parameters and corresponding PSNR values achieved on the EUVP dataset. Our proposed SSD-Net achieves superior performance while maintaining significantly fewer parameters compared to other methods.}
		\label{fig:Fisrt_Fig}
	\end{figure}

	\section{Introduction}
		\IEEEPARstart{T}{he} degradation of underwater image quality primarily stems from complex optical attenuation effects, specifically the dual mechanisms of light scattering and selective absorption. From a physical perspective: (1) The high-density nature of water medium significantly weakens light propagation compared to atmospheric environments, where forward and backward scattering caused by suspended particulates jointly contribute to image blurring; (2) Water molecules exhibit markedly different absorption coefficients across wavelengths—red light suffers attenuation coefficients of 0.6-0.8/m at 1m depth, while blue-green light dominates underwater imaging due to superior penetration capability, directly inducing color distortion through wavelength-selective absorption; (3) Artificial lighting systems often create non-uniform illumination and flare effects underwater, further amplifying imaging noise. The coupling of these degradation factors results in characteristic low SNR, low contrast, and nonlinear color casts in underwater imagery.
		
		As a critical research direction in computational imaging, Underwater Image Enhancement(UIE) employs either physics-model-driven\cite{jaffe1990computer,akkaynak2019sea,chiang2011underwater,berman2020underwater,fu2014retinex} or data-driven approaches\cite{wang2021uiec,10382428,ZHANG2024685,zhao2024wavelet,SMDR-IS,yu2023task,islam2020fast,peng2023u,huang2023contrastive,liu2022twin,Ucolor} to address three core challenges: compensating nonlinear chromatic distortion from light absorption, suppressing spatial blurring caused by scattering, and reconstructing high dynamic range visual perception. Serving as a preprocessing step for high-level vision tasks like underwater image segmentation and object detection.
		
		From an application standpoint, UIE plays pivotal roles in marine ecological monitoring, resource exploration, artifact identification in underwater archaeology, pipeline condition assessment, and visual navigation for Autonomous Underwater Vehicles (AUVs). Particularly in practical engineering scenarios such as aquaculture biomass statistics, shipwreck search-and-rescue operations, and marine mineral resource exploration, high-quality image enhancement can boost operational efficiency and provide reliable perceptual input for underwater intelligent systems.
		
		Prior to the advent of deep learning techniques, underwater image enhancement methods were primarily categorized into physics-based and non-physics-based approaches:
		Physics-based methods rely on the physical mechanisms of underwater light propagation, employing optical attenuation models for image restoration. Representative techniques include the Jaffe-McGlamery imaging model\cite{jaffe1990computer,chen2021underwater,akkaynak2019sea,chiang2011underwater,berman2020underwater,fu2014retinex,chen2019towards}, Underwater Dark Channel Prior (UDCP)\cite{liang2021gudcp,drews2013transmission,peng2017underwater}, and attenuation curve priors\cite{wang2017single,dai2020single}. These methods require estimation of physical parameters such as water attenuation and scattering coefficients. However, due to spatiotemporal variations in underwater environmental parameters, accurate parameter estimation remains challenging in practice. While theoretically grounded with clear physical interpretations, their complex computational processes and strong environmental dependencies limit practical effectiveness.
		
		Non-physics-based methods operate independently of physical imaging models, directly improving visual perception through image processing techniques. Representative approaches encompass histogram equalization\cite{10040560,pizer1987adaptive,hitam2013mixture,li2016underwater}, contrast stretching\cite{li2018polarimetric}, color balance\cite{ancuti2017color,chen2022underwater,bianco2015new,wang2021joint}, and image fusion\cite{zhou2023multi,zhou2022underwater,WWPE}. These methods offer computational efficiency advantages, but the absence of physical constraints often leads to artifacts, over-enhancement, and other irrational corrections, resulting in relatively poor robustness in complex underwater scenarios.
		
		With the advancement of deep learning, CNN-based underwater image enhancement methods have demonstrated significant superiority\cite{wang2021uiec,10382428,ZHANG2024685,zhao2024wavelet,SMDR-IS,yu2023task,islam2020fast,peng2023u,huang2023contrastive,liu2022twin,Ucolor,UWCNN}. These data-driven approaches construct deep neural networks to learn complex nonlinear mappings between degraded and enhanced underwater images from large-scale datasets, exhibiting substantial improvements in feature representation capability and model generalization performance compared to traditional methods.
		
		Deep learning based UIE methods can be primarily categorized into three approaches: Optimization-based methods\cite{WWPE, MMLE}, Generative Adversarial Network (GAN)-based methods\cite{wang2020gan,cong2023pugan,hambarde2021uw,islam2020fast,fabbri2018enhancing} and End-to-end deep network methods\cite{UWCNN,UIEB,Ucolor,SMDR-IS,UIE-PK,CDF-UIE,Phaseformer}. The Sea-Thru\cite{akkaynak2019sea} approach establishes an accurate physical model of underwater imaging, utilizing RGB-D data to estimate critical environmental parameters for color restoration. This method guides the enhancement process through physical parameter inversion. Water-GAN\cite{UIEB} employs an unsupervised learning strategy to address the scarcity of training data. This method leverages GANs to generate realistic degraded underwater images from aerial images and their corresponding depth maps, thereby modeling the underwater degradation process. The synthesized data can subsequently train supervised enhancement networks. As an early representative work, UWCNN\cite{UWCNN} designs a shallow CNN architecture specifically focused on correcting color distortion in underwater images. The network directly learns an end-to-end mapping from degraded to corrected images, primarily optimizing for the prevalent color cast phenomenon in underwater environments.
		
		In the field of image restoration and computer vision, multi-scale feature fusion\cite{xiao2024towards,liu2022deep,ju2020pay,ju2023estimating,ju2023estimating,ju2023gr,zhang2023efficient,yair2018multi,zamir2021multi} has long been regarded as a fundamental technique for extracting image features. The core concept involves combining features with different receptive fields to simultaneously capture both global contextual information and local detail information. From a technical evolution perspective, multi-scale processing has progressed from explicit methods to implicit learning approaches. Traditional computer vision methods explicitly constructed multi-scale representations of input images, such as Gaussian pyramids\cite{li2021multi,lan2015beyond,chang2024iterative} and Laplacian pyramids\cite{burt1987laplacian,paris2011local,chang2024iterative}, to address scale variation challenges. Deep learning approaches, exemplified by U-Net\cite{ronneberger2015u,wang2024ucl,cheng2022light,hashisho2019underwater} and Feature Pyramid Networks (FPN)\cite{lin2017feature,kim2018parallel,mei2023pyramid}, employ neural network architecture designs to implicitly learn multi-scale feature representations.

		Most existing deep learning based UIE methods primarily focus on extracting multi-scale features through various techniques\cite{chen2024uwformer,qi2022sguie,zhang2023hierarchical,chen2021mffn,SMDR-IS,yu2023task,zhang2017underwater,wang2023domain}, under the prevailing assumption that multi-scale features are crucial for improving restoration quality. However, our study reveals that with thorough exploitation of single-scale features, comparable restoration performance can be achieved while significantly reducing computational costs and parameter counts. This leads us to reasonably conjecture that in such tasks, multi-scale features may contain substantial redundant information, whose extraction and processing not only increase computational and model complexity but also contribute minimally to image quality improvement.
		To this end, we propose a novel single-scale feature mining framework—Single-Scale Decomposition Network (SSD-Net). Departing from traditional multi-scale strategies, SSD-Net focuses on fully exploring the potential of single-scale features. To maximize the utilization of single-scale information, SSD-Net incorporates two simple yet effective modules: the Parallel Feature Decomposition Block (PFDB) and the Bidirectional Feature Communication Block (BFCB).
		The PFDB efficiently separates mixed degradation and clear features in underwater images through streamlined convolutions and attention mechanisms. The BFCB employs learnable scaling and a shared attention mechanism to facilitate interaction between degraded and clear features, enabling mutual refinement by extracting residual useful information while suppressing irrelevant components.
		Notably, SSD-Net does not adopt a U-Net-like encoder-decoder structure nor employs pooling or downsampling operations for feature compression. To maintain performance while reducing computational overhead, we extensively utilize depth-wise separable convolution (DSC), achieving an optimal balance between efficiency and effectiveness.

		Our main contributions are summarized as follows:
		\begin{enumerate}
			\item We demonstrate that multi-scale feature extraction is not a necessity in underwater image restoration. When single-scale features are fully explored, comparable performance can be achieved without the complexity introduced by multi-scale designs, leading to significant reductions in computational cost and parameter count.
			
			\item To effectively exploit the potential of single-scale features, we propose the Single-Scale Decomposition Network (SSD-Net), which leverages the synergy between two key components: the Parallel Feature Decomposition Block (PFDB) and the Bidirectional Feature Communication Block (BFCB). Together, these modules enable efficient feature extraction and refinement.
			
			\item Extensive experiments validate our findings. A series of comparative studies show that our method outperforms current state-of-the-art (SOTA) approaches, and ablation studies further confirm the effectiveness and rationality of our proposed design.
		\end{enumerate}  	
	
	\section{Related Works}
	Before deep learning became widely adopted for image enhancement tasks \cite{bui2024edge,chen2023remote,kang2024act,ibrahim2025hyda,patnaik2024two,shang2022spectral}, traditional underwater image enhancement methods primarily relied on either understanding light propagation characteristics in water or employing general image processing techniques. These approaches can be broadly categorized into two groups: model-based methods\cite{jaffe1990computer,akkaynak2019sea,chiang2011underwater,berman2020underwater,fu2014retinex,peng2017underwater} and image-based methods\cite{10040560,pizer1987adaptive,hitam2013mixture,li2016underwater,li2018polarimetric,ancuti2017color,chen2022underwater,bianco2015new,wang2021joint,zhou2023multi,zhou2022underwater,WWPE}.
	model-based methods attempt to reverse the image degradation process by constructing optical models of underwater image formation such as the Jaffe-McGlamery model\cite{jaffe1990computer} and estimating their parameters. These approaches typically involve estimating transmission maps and background light to approximately reconstruct the image's original appearance in a water-free environment. They heavily rely on accurate estimation of environmental parameters (e.g., water depth, turbidity, background light). 
	Akkaynak et al.\cite{akkaynak2019sea} proposed Sea-thru, a physics-accurate underwater image color-restoration technique that overturns the assumptions of conventional atmospheric scattering models by introducing underwater light-attenuation and backscatter coefficients. This physics-driven approach fundamentally resolves underwater color distortion, yet its reliance on absolute depth acquired via Structure-from-Motion or stereo vision constrains its broader applicability.
	
	However, robustly estimating these parameters in complex, variable, and unknown real-world underwater environments is inherently challenging. For example, methods based on the Underwater Dark Channel Prior (UDCP)\cite{liang2021gudcp} or Underwater Light Attenuation Prior (ULAP)\cite{song2018rapid} typically depend on scene-specific priors to estimate transmission or depth. When these underlying assumptions fail or parameter estimation is inaccurate, enhancement results not only underperform but may also introduce artifacts or unnatural visual appearances.
	By jointly exploiting image blurriness and the red-light absorption characteristic for depth estimation, Peng et al.\cite{peng2017underwater} proposed a method that overcomes the failure of conventional DCP and Maximum Intensity Prior (MIP) under complex underwater lighting conditions, thereby addressing the fundamental issues of color distortion and low contrast in underwater images.
	Non-physics-based methods, on the other hand, operate without explicit imaging models, instead directly manipulating pixel values to improve visual quality. Common techniques include histogram equalization, contrast stretching, color correction in various color spaces, Retinex theory, and various image fusion strategies. For instance, Ancuti et al.\cite{ancuti2012enhancing,ancuti2017color} explicitly adopted fusion strategies by combining color-corrected and contrast-enhanced image versions to improve multiple quality dimensions simultaneously. Similarly, Li et al.\cite{8792133} jointly addressed dehazing and color correction. These works demonstrate that single enhancement operations struggle to comprehensively improve image quality, whereas fusion or multi-stage strategies enable targeted optimization of different image attributes before combining them through reasonable fusion mechanisms to produce results with superior overall perceptual quality.
	
	The above mentioned factors constitutes a fundamental bottleneck for traditional methods, leading to inconsistent performance across different scenarios. Consequently, deep learning approaches\cite{UIEB}, which capable of implicitly learning complex degradation patterns and restoration strategies from data, have gradually become the mainstream in underwater image enhancement research.
	
	Deep learning methods in underwater image enhancement\cite{UWCNN,Ucolor,semi-uir,SMDR-IS,UIE-PK,CDF-UIE,Phaseformer,DiffWater,zhou2023ugif} demonstrate a strong trend towards more sophisticated network architectures—such as U-Net\cite{Ronneberger2015unet}, multi-branch networks\cite{srivastava2015highway,he2016deep,abdi2016multi,li2019selective}, attention mechanisms\cite{mnih2014recurrent,bahdanau2014neural,xu2015show}, Transformer\cite{vaswani2017attention} and more refined learning strategies, including GANs\cite{goodfellow2014generative}, reinforcement learning, and self-supervised learning. This trend is driven by the need to capture complex spatial and color relationships within images, enhance the realism of generated outputs, and reduce reliance on perfectly paired training data—a major bottleneck in this field. For instance, the introduction of GANs\cite{UIEB} aims to produce more realistic images and enable unsupervised learning. Architectures like U-Net are favored for image-to-image translation tasks due to their multi-scale feature processing capabilities. Attention mechanism and Transformer are employed to capture long-range dependencies critical for maintaining global image consistency. Some approaches explicitly process images across multiple color spaces\cite{lin2023tcrn,wang2021uiec}, while others attempt to fuse physical priors or imaging model features\cite{liu2024coc}. Future advances are likely to lie in developing hybrid models that synergize traditional domain knowledge with the powerful capabilities of modern machine learning, thereby producing more robust, efficient, and interpretable underwater image enhancement solutions. 
	
	\begin{figure}[b]
		\centering
		\includegraphics[width=0.48\textwidth]{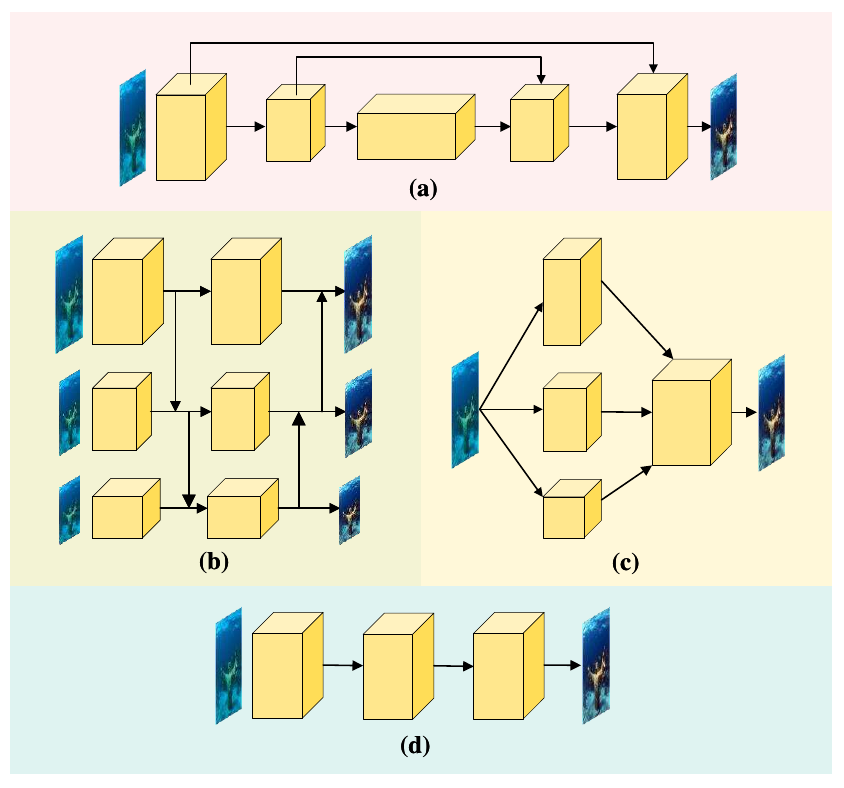}
		\caption{Comparison of several common image restoration network architectures: (a). U-Net and its variants, (b). Progressive Multi-Resolution, (c). Parallel Multi-Branch, and (d). Single-Scale Network.}
		\label{fig:architecture}
	\end{figure}
	
	
	The aforementioned deep learning-based methods almost universally employ multi-scale features for underwater image feature extraction, based on the premise that multi-scale features are generally beneficial for image restoration under various conditions. Some explicitly utilize feature pyramids\cite{lin2017feature,ghiasi2019fpn,jiang2023five}, others adopt multi-branch parallel multi-scale convolutional structures, while many leverage U-Net architectures and their variants. However, an excessive reliance on multi-scale features may introduce redundancy and impose computational burdens. Our experiments reveal that although using multi-scale features achieves excellent restoration results, it also generates a substantial amount of redundant features, and increasing the number of features does not necessarily lead to better performance. Further experiments demonstrate that by fully exploiting the information contained in single-scale features, comparable restoration quality can be attained without employing multi-scale extraction and fusion strategies, while significantly reducing computational resource consumption.
	
	
	\section{Proposed Method}
	In this section, we will present the details of our proposed SSD-Net, which is designed to enhance underwater images using only single-scale features. First, we will introduce the overall architecture and workflow of the network. Then, we will explain the two key modules we have designed: the Parallel Feature Decomposition Block (PFDB) and the Bidirectional Feature Communication Block (BFCB). Finally, we will provide a detailed explanation of the components of the loss function.

	\subsection{Motivation}\label{Motivation}
	As previously mentioned, in recent years, researchers designing underwater image enhancement methods have often considered extracting multi-scale features to assist in the enhancement process. Multi-scale techniques are widely applied in many computer vision tasks, especially in image segmentation and image detection. Indeed, previous methods utilizing multi-scale features for underwater image enhancement have achieved remarkable results. The frameworks based on multi-scale feature extraction can generally be categorized into three main types, as shown in Fig. \ref{fig:architecture}: (a) U-Net and its variants, which use continuous downsampling and upsampling in a ``contraction-expansion" path to capture hierarchical features and directly pass high-resolution details to the decoding end through skip connections. (b) Progressive Multi-Resolution, where the core idea is to process the same image at different resolutions in a progressive manner—coarse corrections are made at low resolution first, followed by high-resolution fine details, with explicit feature transmission or residual prediction at each level. (c) Parallel multi-branch, where the core idea is to parallelize several convolution or attention branches within the same layer, each using different receptive fields or kernel sizes. The results of the branches are then concatenated or weighted to form a multi-scale description. However, these three approaches also have obvious drawbacks, primarily in terms of large parameter and computational costs, as well as feature redundancy. Taking U-Net and its variants as an example, such networks perform both downsampling and upsampling on the same hierarchical features during encoding and decoding, and the computational and parameter costs increase rapidly with the depth of the network. Skip connections typically use channel concatenation to directly pass high-resolution features, which not only increases memory usage but also introduces a significant amount of low-level features that overlap highly with the already encoded high-level information, leading to feature redundancy and noise accumulation. Therefore, channel attention or gating mechanisms are often introduced to suppress these issues. Progressive Multi-Resolution processes the same image through multi-stage serial sub-networks. Although the initial stage provides some computational advantage at low resolution, subsequent stages revert to high-resolution processing, leading to exponential growth in overall computation and parameter numbers. Due to the lack of weight sharing or explicit feature distillation across levels, edge and texture information learned in early stages is repeatedly encoded in later stages, resulting in redundant computations and features across levels. The parallel multi-branch architecture connects different convolution kernel sizes or attention branches within the same layer, and the computational and memory overhead increases linearly with the number of branches. To enhance performance, it is often necessary to increase the number of branches or expand the channel width, thus magnifying the computational cost and parameter scale. Due to significant overlap in receptive fields across branches, the output features are semantically highly related, which causes a considerable expansion in channel dimensions and a reduction in feature information. To reduce redundant features and control computational costs, the model often needs to employ 1×1 convolutions or channel attention to compress the outputs of branches. As shown in Fig. \ref{fig:architecture}(d), the seemingly simple single-scale feature network, after being reconstructed with large kernels, frequency domain modules, SSM, dynamic convolutions, and other techniques, can effectively overcome the limited receptive fields and weak semantic representations inherent in conventional single-scale architectures. The SSD-Net proposed in this paper is an improvement on single-scale feature networks. By designing efficient feature decomposition and feature interaction refinement modules, we achieve underwater image enhancement without relying on multi-scale features, and our extensive experiments demonstrate the efficiency of our method.

	\begin{figure*}[t]
		\centering
		\includegraphics[width=0.95\textwidth]{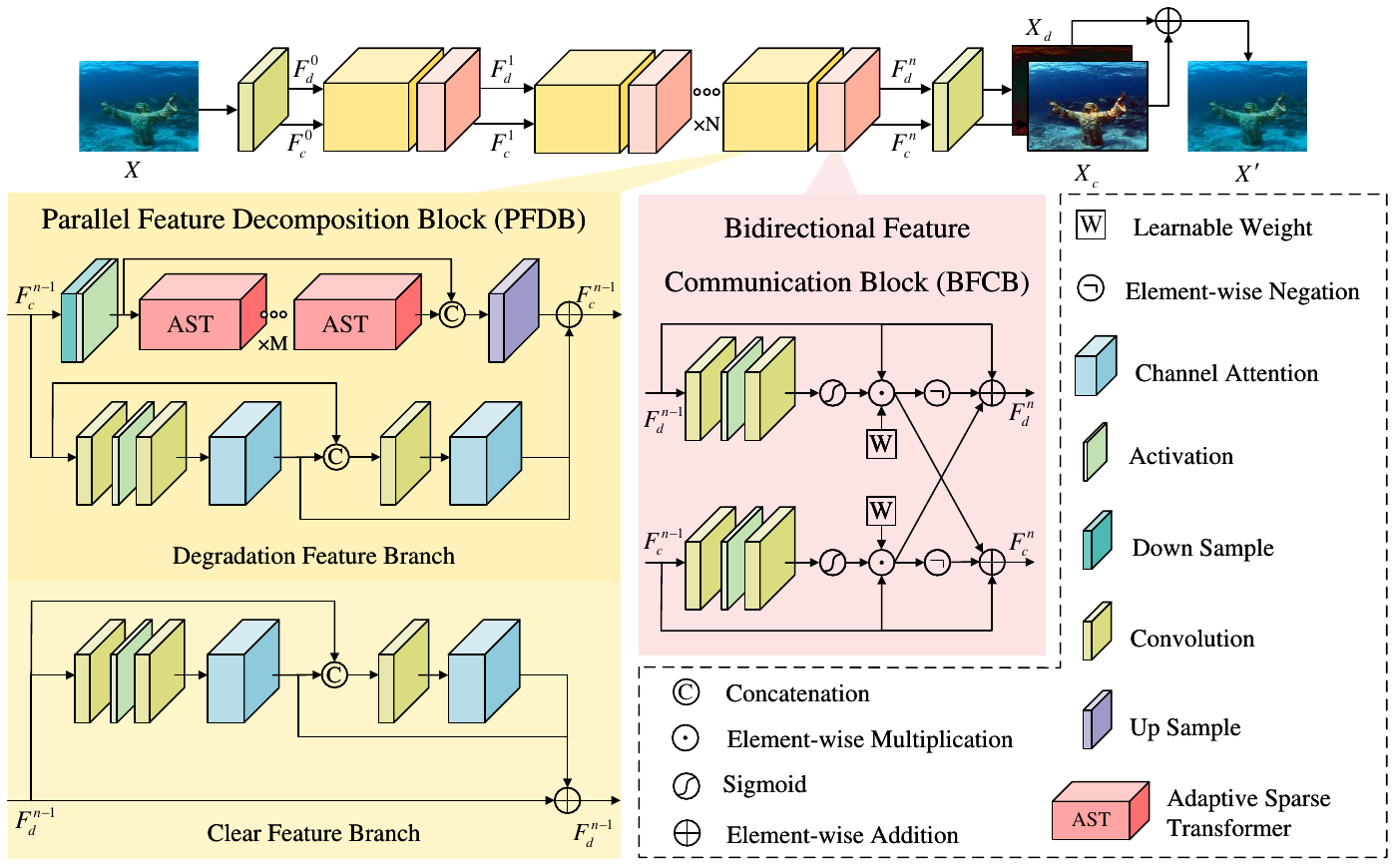}
		\caption{An overview of the proposed SSD-Net. The network is designed as a single-scale architecture, consisting of cascaded PFDB and BFCB modules. It takes an underwater image $X$ as input and maps it to the feature space, where the degradation features and clear features are separated, interacted, and refined. Finally, the network extracts the clear image $X_c$ and the degradation residual map $X_d$, and adds them element-wise to obtain the predicted underwater image $X'$, which then participates in the global network parameter optimization.}
		\label{fig:overview}
	\end{figure*}
	
	\subsection{Overall Architecture}
	Fig. \ref{fig:overview} illustrates the overall workflow of SSD-Net, which comprises a convolutional feature-embedding unit, cascaded PFDB and BFCB modules, and a convolutional reconstruction module. According to this workflow, the feature-embedding unit first maps the input underwater image $X \in \mathbb{R}^{H \times W \times C}$ into the feature space, where $H$, $W$, and $C$ denote the height, width, and number of channels, respectively. This operation can be formally expressed as
	
	\begin{equation}
		F^{0}_d, F^{0}_c = \theta(X, \phi),
	\end{equation}
	where $\theta(\cdot)$ denotes the convolution operation and $\phi$ represents the learnable parameters. Here, $F_d$ denotes the clear features that facilitate clean-image restoration, while $F_c$ represents the degradation features that help predict the degradation factors in the image. Next, the cascaded PFDB and BFCB modules take $F_d$ and $F_c$ as input and output the refined features $F^{n}_d$ and $F^{n}_c$:
	
	\begin{equation}
		\begin{split}
			F^{1}_d, F^{1}_c &= BFCB_1(PFDB_1(F^{0}_d, F^{0}_c)), \\
			&\;\;\vdots \qquad  \\
			F^{n}_d, F^{n}_c &= BFCB_n(PFDB_n(F^{n-1}_d, F^{n-1}_c)).
		\end{split}
	\end{equation}
	Finally, $F^{n}_d$ and $F^{n}_c$ are fed into the reconstruction module to generate the clear image $X^c \in \mathbb{R}^{H \times W \times C}$ and the degradation residual map $X^d \in \mathbb{R}^{H \times W \times C}$:
	
	\begin{equation}
		X^c, X^d = \varphi((F^{n}_d, F^{n}_c), \mu),
	\end{equation}
	where $\varphi(\cdot)$ denotes the last reconstruction module and $\mu$ represents its learnable parameters.
	The degradation residual map is the predicted difference between the reference image and the input image, capturing the degradation part of the underwater image. Finally, the network reconstructs the underwater image $X'$ by adding the predicted clear image $X_c$ and the degradation residual map $X_d$ element-wise. Training is supervised by computing the loss between $X_c$ and the reference image $Y \in \mathbb{R}^{H \times W \times C}$, as well as the loss between $X'$ and the input image $X$.
	
	\subsection{Parallel Feature Decomposition Block}

\begin{figure}[htbt]
		\centering
		\includegraphics[width=0.45\textwidth]{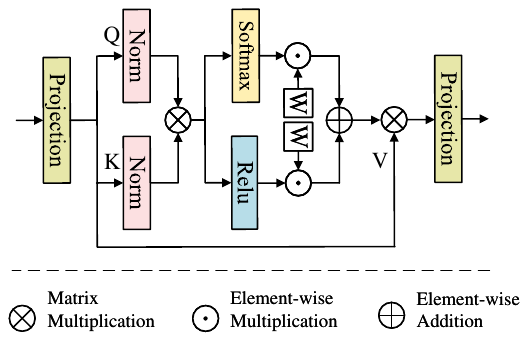}
		\caption{illustrates the proposed Adaptive Sparse Attention mechanism. The key idea is to divide the Transformer's self-attention into dense and sparse branches and dynamically fuse them to achieve a more precise attention representation..}
		\label{fig:transformer}
	\end{figure}

	
	To achieve accurate inference with a dual-branch decoupled network, we combine the strengths of Transformer and CNN. For the complex residual degradation components in underwater images, the global modeling capability of Transformer effectively captures degradation-related information. In the traditional self-attention mechanism, similarity scores are first obtained by the scaled dot product between the Query and Key, followed by a Softmax operation. Softmax amplifies high scores via the exponential function $e^{x_i}$—thereby highlighting keys highly correlated with the query—while suppressing lower scores and directing the model’s focus to salient information. For example, if a key $k_i$ is highly related to the query $q_i$, its Softmax weight will be substantially higher than those of other keys. However, because attention scores are computed over all tokens, this dense scheme often introduces irrelevant regions, causing the attention maps to aggregate redundant information and hindering the extraction of valid features. To introduce sparsity into standard self-attention, as shown in Fig. \ref{fig:transformer}, we replace Softmax with a ReLU activation after computing the similarity scores, directly filtering out negative values. To prevent excessive sparsity introduced by ReLU, we retain a parallel Softmax branch. The attention scores from the two branches are then adaptively fused via learnable weights, which suppress noise and redundancy while preserving informative features to the greatest extent. The computation of adaptive sparse attention can be formulated as:
	\begin{equation}
		Q, K, V = {split}({DWConv}_{3\times3}({Conv}_{1\times1}(X))),
	\end{equation}
	We first expand the channel dimension of the feature map $X$ via a linear mapping $Conv_{1\times1}(\cdot)$ and then employ a $3\times3$ DSC $DWConv_{3\times3}$ to capture local spatial information. Then incorporation of depth-wise separable convolution (DSC) substantially enhances spatial modeling ability while introducing minimal additional parameters.
	\begin{equation}
		A = (norm(Q) \cdot norm(K)^T) \odot T,
	\end{equation}

	\begin{equation}
		\begin{split}
			&\quad \alpha_{\text{dense}} = {softmax}(A), \\
			&\quad \alpha_{\text{sparse}} = \frac{{ReLU}(A)}{\sum {ReLU}(A) + \epsilon}, \\
			&\quad \alpha = w_{\text{dense}} \cdot \alpha_{\text{dense}} + w_{\text{sparse}} \cdot \alpha_{\text{sparse}}, \\
		\end{split}
	\end{equation}

	\begin{equation}
		A = \alpha \cdot V.
	\end{equation}
	Here, $norm(\cdot)$ denotes the normalization operation, $(\cdot)^{\mathrm{T}}$ denotes the transpose, $\odot$ represents the dot-product operation, $\epsilon$ is a small constant added to avoid division by zero, and $w_{\text{dense}}$ and $w_{\text{sparse}}$ are learnable weight parameters.
	
	For the clear content inherent in underwater images, we adopt a lightweight CNN structure augmented with two layers of channel attention to focus on local feature representation. As shown in Fig. \ref{fig:overview}, the PFDB feeds the input features $F^{n-1}_d$ and $F^{n-1}_c$ into a Transformer-driven Degradation Feature Branch and a CNN-driven Clear Feature Branch, respectively. Because the entire framework operates on single-scale features, we sandwich the Transformer blocks with down-sampling and up-sampling operations to reduce computational cost; to compensate for the Transformer's limited local modeling, this branch also incorporates the same dual channel-attention module used in the CNN branch. It is worth noting that, to further reduce computational overhead, depthwise separable convolutions are employed throughout all channel-attention modules. Taking Fig. \ref{fig:overview} as an example, the overall computation of the PFDB can be formulated as
	
	{\footnotesize
		\begin{gather}
			\widehat{F_d^{n - 1}} = CA([F_d^{n - 1}, CA(F_d^{n - 1})]) + [AST(F_d^{n - 1} \downarrow), F_d^{n - 1} \downarrow] \uparrow, \\
			\widehat{F_c^{n - 1}} = CA([F_c^{n - 1}, CA(F_c^{n - 1})]).
		\end{gather}
	}
	where $AST(\cdot)$ denotes stacked Adaptive Sparse Transformer blocks, $\downarrow$ and $\uparrow$ represent down-sampling and up-sampling operations, respectively, $[\cdot,\cdot]$ denotes feature concatenation, and $CA(\cdot)$ represents the channel-attention module with DSC.
	
	\subsection{Bidirectional Feature Communication Block}
	After passing through the PFDB, the features of the two branches already possess strong representational power. To further refine them, we introduce the BFCB. As mentioned earlier, both $F_d$ and $F_c$ are embedded by the initial convolutional layers; therefore, at the coarse representation stage they inherently contain considerable mutual correlation. Even after the targeted extraction performed by the PFDB, these correlations cannot be fully removed. BFCB is designed to mine the useful information implicitly contained in the counterpart while eliminating redundant information within its own branch, thus decoupling and fusing the two feature sets and enabling bidirectional interaction. As shown in Fig. \ref{fig:overview}, this refinement module adopts a symmetric structure: $\widehat{F_d^{n-1}}$ and $\widehat{F_c^{n-1}}$ are first fed into two successive $1\times1$ convolutions with a ReLU activation in between, and a Sigmoid function is applied to rescale the outputs to $[-1,1]$ to obtain the fusion weights:

	{\footnotesize
	\begin{equation}
		\begin{split}
			&\quad G_{c \rightarrow d}^{n-1} = \sigma\left({Conv}_{1\times1}\left({ReLU}\left({Conv}_{1\times1}(\widehat{F_d^{n - 1}})\right)\right)\right), \\
			&\quad G_{d \rightarrow c}^{n-1} = \sigma\left({Conv}_{1\times1}\left({ReLU}\left({Conv}_{1\times1}(\widehat{F_c^{n - 1}})\right)\right)\right). \\
		\end{split}
	\end{equation}
	}
	Here, $\sigma(\cdot)$ denotes the Sigmoid function. Next, the features are multiplied by their corresponding fused weights and learnable scaling factors to obtain the coupled residuals:

	\begin{equation}
		\begin{split}
			&\quad {res}_{c \rightarrow d} = \widehat{F_c^{n - 1}} \cdot G_{c \rightarrow d}^{n-1} \cdot w_{c \rightarrow d}^{n - 1}, \\
			&\quad {res}_{d \rightarrow c} = \widehat{F_d^{n - 1}} \cdot G_{d \rightarrow d}^{n-1} \cdot w_{d \rightarrow c}^{n - 1}. \\
		\end{split}
	\end{equation}
	Here, ${res}_{c \rightarrow d}$ denotes the degradation residual information extracted from the clear features, and ${res}_{d \rightarrow c}$ represents the reverse direction. We then subtract the irrelevant components from each feature and add the coupled residuals from the opposite branch to obtain a more refined feature representation:
	\begin{equation}
		\begin{split}
			&\quad F^{n}_d = \widehat{F_d^{n - 1}} - {res}_{d \rightarrow c} + {res}_{c \rightarrow d}, \\
			&\quad F^{n}_c = \widehat{F_c^{n - 1}} - {res}_{c \rightarrow d} + {res}_{d \rightarrow c}. \\
		\end{split}
	\end{equation}
	Learnable scaling parameters allow the network to adaptively modulate the strength of information flow, enhancing flexibility. Residual connections mitigate gradient-vanishing while preserving the original feature information, thereby improving training stability and final performance. Leveraging a sophisticated fusion structure and adaptive weighting, the BFCB enables efficient, flexible bidirectional feature interaction: it captures richer feature relationships and dynamically adjusts information flow according to task requirements, resulting in overall performance gains.

	\subsection{Loss Function}\label{Loss Function}

	As introduced earlier, our model produces two outputs: the predicted clear image $X_c$ and the degradation residual $X_d$. The reconstructed underwater image $X'$ is obtained by summing $X_c$ and $X_d$. To ensure the robustness of the model, it is essential to define an appropriate loss function that quantifies the consistency between the model predictions and the reference data during training.
	
	To evaluate the quality of the predicted clear image and the reconstructed degraded image, we incorporate a structural similarity-based loss to guarantee sufficient similarity between the predicted and reference images.
	
	The structural similarity index (SSIM) is a widely-used perceptual metric that measures the similarity between two images in terms of structure, luminance, and contrast. SSIM accounts for local structural information and aligns better with the human visual system. The SSIM-based loss function is defined as:
	\begin{equation}
		\mathcal{L}_{ssim}=1 - {\left( {\frac{{2{\mu _e}{\mu _r} + {c_1}}}{{\mu _e^2 + \mu _r^2 + {c_1}}}} \right)^\alpha }{\left( {\frac{{2{\sigma _{er}} + {c_2}}}{{\sigma _e^2 + \sigma _r^2 + {c_2}}}} \right)^\beta },
	\end{equation}
	where $\mu _e$ and $\mu _r$ denote the mean values of the predicted and reference images, respectively. $\sigma _e$ and $\sigma _r$ are the corresponding standard deviations, and $\sigma _{er}$ denotes the covariance between them. $\alpha$ and $\beta$ control the relative importance of the two terms, and $c_1$ and $c_2$ are constants introduced to stabilize the division.
	
	In addition to the SSIM loss, we employ the L1 loss as the reconstruction loss. L1 loss is commonly used in UIE tasks and is formulated as:
	\begin{equation}
		\mathcal{L}_{rec}=\frac{1}{N}\sum_{i=1}^{N}{{\left|y_i' - y_i\right|}_1},
	\end{equation}
	where $N$ is the number of pixels in the image, and $i$ denotes the pixel index. In UIE tasks, L1 loss minimizes pixel-wise errors between the generated image and the ground truth. Compared to L2 loss, L1 is more robust to outliers and better handles noise, avoiding over-smoothing in the output.
	
	To further guide the network learning process, we apply the same loss functions to the reconstructed underwater image $X'$. The final loss function is constructed as follows:
	
	\begin{equation}
		\begin{split}
			\mathcal{L}_{total} = \mathcal{L}_{ssim}(X^c, Y) + \mathcal{L}_{1}(X^c, Y) + \\
			 \alpha\mathcal{L}_{ssim}(X', X) + \beta\mathcal{L}_{1}(X', X)
		\end{split}
	\end{equation}
	where $\alpha$ and $\beta$ representing the hyperparameters associated with each individual loss component.

	\begin{table*}[htbp]
		\renewcommand{\arraystretch}{1.3}
		\centering
		\caption{{Quantitative results of UIE on the UIEB, EUVP and UIQS datasets as well as computational complexity, we use SSIM↑, PSNR↑, LPIPS↓ and MSE ($\times10^3$)↓ for paired datasets and UIQM↑, UCIQE↑ for unpaired datasets, where ↑ larger values mean better results and ↓ denotes that smaller values mean better results. The \textbf{best} and \underline{second} results are shown in boldface and underscore.}}
		\resizebox{\linewidth}{!}{
				\begin{tabular}{cccccccccccc}
			\hline\noalign{\vskip 2pt}
			\multicolumn{1}{c|}{\multirow{2}[2]{*}{Dataset}} & \multicolumn{1}{c|}{\multirow{2}[2]{*}{Metric}} & \multicolumn{1}{c|}{UWCNN} & \multicolumn{1}{c|}{Ucolor} & \multicolumn{1}{c|}{MMLE} & \multicolumn{1}{c|}{Semi-UIR} & \multicolumn{1}{c|}{WWPE} & \multicolumn{1}{c|}{SMDR-IS} & \multicolumn{1}{c|}{UIE-PK} & \multicolumn{1}{c|}{CDF-Net} & \multicolumn{1}{c|}{Phaseformer} & \multirow{2}[2]{*}{Ours} \bigstrut[t]\\
			\multicolumn{1}{c|}{} & \multicolumn{1}{c|}{} & \multicolumn{1}{c|}{20'PR} & \multicolumn{1}{c|}{21'TIP} & \multicolumn{1}{c|}{22'TIP} & \multicolumn{1}{c|}{23'CVPR} & \multicolumn{1}{c|}{23'TCSVT} & \multicolumn{1}{c|}{24'AAAI} & \multicolumn{1}{c|}{25'ICASSP} & \multicolumn{1}{c|}{25'TGRS} & \multicolumn{1}{c|}{25'WACV} &  \bigstrut[b]\\\noalign{\vskip 2pt}
			\hline
			\hline
			\multirow{4}[2]{*}{UIEB} & SSIM↑ & 0.652 & 0.888 & 0.755 & 0.884 & 0.767 & 0.903 & 0.893 & 0.831 & \underline{0.909} & \textbf{0.924} \bigstrut[t]\\
			& PSNR↑ & 14.08 & 22.56 & 17.02 & 22.48 & 17.97 & 22.51 & 22.52 & 21.32 & \underline{23.02} & \textbf{24.90} \\
			& LPIPS↓ & 0.262 & 0.077 & 0.240 & 0.090 & 0.186 & 0.080 & 0.098 & 0.202 & \underline{0.073} & \textbf{0.062} \\
			& MSE↓  & 3.234 & \underline{0.466} & 1.650 & 0.548 & 1.289 & 0.549 & 0.535 & 0.599 & 0.557 & \textbf{0.440} \bigstrut[b]\\
			\hline
			\multirow{4}[2]{*}{EUVP} & SSIM ↑ & 0.719 & 0.828 & 0.643 & 0.797 & 0.656 & 0.894 & 0.892 & 0.881 & \underline{0.899} & \textbf{0.917} \bigstrut[t]\\
			& PSNR↑ & 17.22 & 20.58 & 15.01 & 19.13 & 15.89 & \underline{28.00} & 26.69 & 26.58 & 27.40 & \textbf{29.84} \\
			& LPIPS↓ & 0.277 & 0.244 & 0.253 & 0.238 & 0.287 & \underline{0.078} & 0.121 & 0.173 & 0.082 & \textbf{0.058} \\
			& MSE↓  & 1.578 & 0.672 & 2.507 & 0.934 & 2.011 & \underline{0.125} & 0.164 & 0.174 & 0.142 & \textbf{0.083} \bigstrut[b]\\
			\hline
			\multirow{2}[2]{*}{UIEB60} & UIQM↑ & 3.068 & 3.448 & 3.211 & 3.543 & \underline{3.681} & 3.597 & 3.654 & 3.510 & 3.605 & \textbf{3.698} \bigstrut[t]\\
			& UCIQE↑ & 0.482 & 0.538 & 0.592 & 0.575 & \underline{0.594} & 0.576 & 0.583 & 0.573 & 0.576 & \textbf{0.598} \bigstrut[b]\\
			\hline
			\multirow{2}[2]{*}{U45} & UIQM↑ & 4.680 & 5.072 & 4.831 & 5.100 & 4.818 & \textbf{6.869} & 5.139 & 4.958 & 5.086 & 5.099 \bigstrut[t]\\
			& UCIQE↑ & 0.471 & 0.564 & 0.598 & 0.600 & 0.599 & 0.606 & 0.593 & \underline{0.609} & 0.605 & \textbf{0.611} \bigstrut[b]\\
			\hline
			\multirow{2}[2]{*}{UCCS} & UIQM↑ & 3.155 & 3.603 & 4.140 & 3.846 & \underline{4.232} & 3.725 & 3.896 & 3.559 & \textbf{4.535} & 3.869 \bigstrut[t]\\
			& UCIQE↑ & 0.467 & 0.530 & 0.579 & 0.567 & \textbf{0.593} & 0.529 & 0.538 & 0.565 & 0.544 & \underline{0.582} \bigstrut[b]\\
			\hline
			\multicolumn{2}{c}{Params} & 0.04M & 157.40M & \textbackslash{} & 65.60M & \textbackslash{} & 12.56M & 1.84M & 15.90M & 1.779M & 1.03 M \bigstrut\\
			\hline
		\end{tabular}%
		}
		\label{tab:SOTA}%
	\end{table*}%
	
	\section{Experiments}
	In this section, we first describe the experimental settings. Then, we compare the performance of our method against several SOTA approaches on both reference-based and non-reference UIE datasets. Finally, we conduct ablation studies to evaluate the effectiveness of each component in our model.
	\subsection{Experimental Sections}
	\subsubsection{\textbf{Implementation Details}}
	Our SSD model is implemented using the PyTorch framework and trained on a single NVIDIA RTX 3090 GPU. We adopt the Adam optimizer with a batch size of 4. The initial learning rate is set to $3 \times 10^{-4}$ and gradually decays to $3 \times 10^{-5}$. The total number of training epochs is 500. In the model architecture, four pairs of PFDB and PFCB modules are cascaded, and each PFDB contains four AST modules. The parameters $\alpha$ and $\beta$ in Eq.(15) are both empirically set to 0.2.
	
	\subsubsection{\textbf{Comparison Methods}}
	We conduct comparative experiments against nine publicly available UIE methods from recent years, including UWCNN\cite{UWCNN}, Ucolor\cite{Ucolor}, MMLE\cite{MMLE}, Semi-UIE\cite{semi-uir}, WWPE\cite{WWPE}, SMDR-IS\cite{SMDR-IS}, UIE-PK\cite{UIE-PK}, CDF-Net\cite{CDF-UIE}, and Phaseformer\cite{Phaseformer}. All corresponding papers have been published in well-recognized journals or conferences. Among these, MMLE and WWPE are traditional approaches, while the others are based on deep learning. For fair comparison, each method is tr ained and tested following the experimental protocols described in their respective papers.
	
	\subsubsection{\textbf{Evaluation Datasets}}
	We employ two reference-based datasets widely used in UIE tasks: UIEB (Underwater Image Enhancement Benchmark)\cite{UIEB} and EUVP (Enhancing Underwater Visual Perception)\cite{islam2020fast}. UIEB contains 890 raw underwater images paired with corresponding high-quality reference images. We randomly select 790 images for training and the remaining 100 for testing. For EUVP, we adopt the Underwater Scenes subset, which includes 2,185 paired images. Among them, 1,985 pairs are randomly selected for training, and 200 pairs are used for testing. 
	
	For unpaired datasets, we adopt the Challenge-60 subset from UIEB and we denote it as UIEB60, the U45\cite{U45} dataset, and the UCCS (Underwater Color Cast Set) subset from RUIE\cite{RUIE} (Real-world Underwater Image Enhancement). The Challenge-60 subset contains 60 particularly challenging no-reference images spanning six difficult scenarios—blue tone, green tone, yellow tone, turbid colorless, turbid colored, and low-light—and is widely used to evaluate model generalization in extreme environments . U45 consists of 45 real-world underwater images at a resolution of 256×256, covering various degradation types such as color cast, low contrast, and haze UCCS includes approximately 300 images categorized into blue, green, and bluish-green casts, specifically designed to assess underwater color correction performance.
	
	\subsubsection{\textbf{Evaluation Metrics}}
	we employ a variety of evaluation metrics to comprehensively assess model performance. For paired datasets, we use SSIM, PSNR, LPIPS\cite{LPIPS}, and MSE ($\times10^3$). Specifically, SSIM measures structural similarity, providing an effective assessment of perceptual image quality; PSNR quantifies pixel-level reconstruction errors, with higher values indicating more accurate restoration; LPIPS leverages deep features to evaluate perceptual quality, capturing fine details and textures; MSE computes the average squared pixel-wise difference, offering a simple yet structure-agnostic metric. For unpaired datasets, we adopt no-reference metrics, namely UIQM and UCIQE. UIQM evaluates image quality based on luminance, contrast, and structural attributes, making it suitable for general-purpose quality assessment. UCIQE, designed specifically for underwater imagery, effectively measures clarity, color fidelity, and contrast. By integrating multiple evaluation metrics, we can comprehensively assess enhancement performance across various aspects and conditions, ensuring that the evaluation reflects the true capabilities of the model and that comparisons are conducted fairly.
	
	\subsection{Comparison with SOTA Methods}
	\subsubsection{\textbf{Quantitative Comparison}}
	We evaluate our method on paired datasets UIEB and EUVP, with results summarized in TABLE \ref{tab:SOTA}. On both datasets, our approach significantly outperforms existing SOTA methods across all metrics. Specifically, on UIEB our method achieves 1.65\% and 8.17\% improvements in SSIM and PSNR, respectively, compared to the second-best method, Phaseformer. On EUVP, we observe a 2\% increase in SSIM and a 6.57\% gain in PSNR over the second-best method, SMDR‑IS. These results demonstrate the superiority of our approach on reference-based benchmarks.
	
	We further evaluate our approach on the no-reference datasets UIEB60, U45, and UCCS, with results shown in the bottom half of TABLE \ref{tab:SOTA}. Our method consistently outperforms competing techniques in all four metrics on UIEB60 and U45. Although it performs slightly lower on UCCS, it still achieves competitive results. Overall, across these three unpaired datasets, our method demonstrates strong performance in no-reference scenarios.
	
	Finally, we present a comparison of model parameter sizes in the last row of  TABLE \ref{tab:SOTA}. Our model ranks second lightest among all methods, although UWCNN has fewer parameters, its significantly inferior performance highlights that simply reducing parameter count is insufficient without effective feature extraction strategies, as demonstrated by our proposed method. In contrast, compared with other recent competitive approaches, our model achieves superior quantitative results while maintaining a remarkably low parameter footprint, further underscoring the efficiency–effectiveness balance enabled by our design.
	\begin{figure*}[h]
		\centering
		\includegraphics[width=0.98\textwidth]{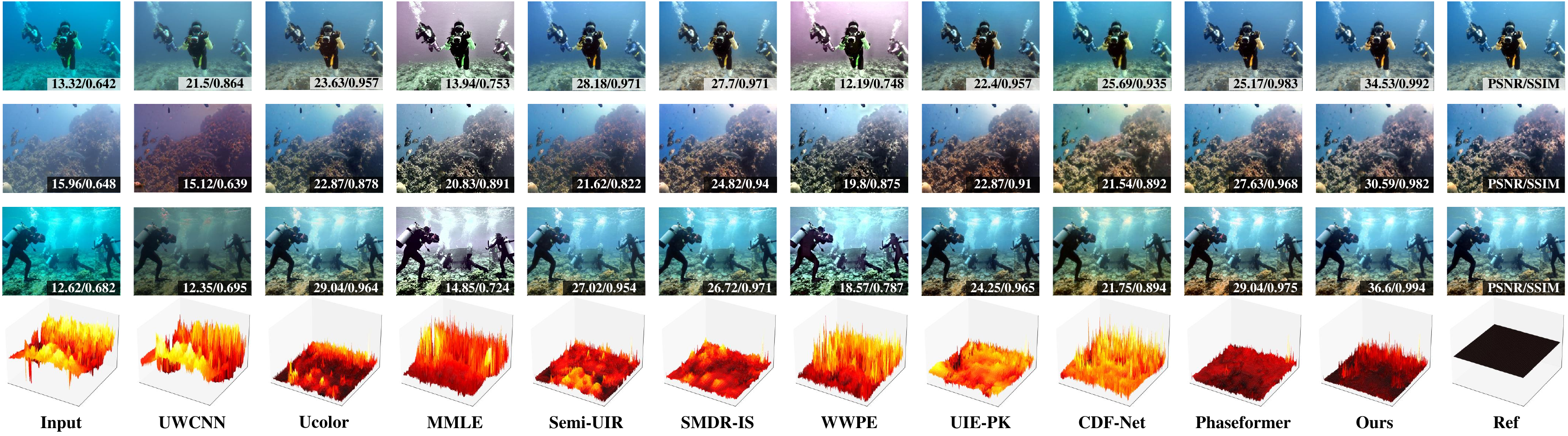}
		\caption{Qualitative comparison on the UIEB dataset. Please zoom in for better visualization of fine details. The bottom-right corner of each image shows the corresponding PSNR and SSIM values. The last row presents the error maps between the third-row results and the reference image.}
		\label{fig:UIEB}
	\end{figure*}
	
	\begin{figure*}[h]
		\centering
		\includegraphics[width=1\textwidth]{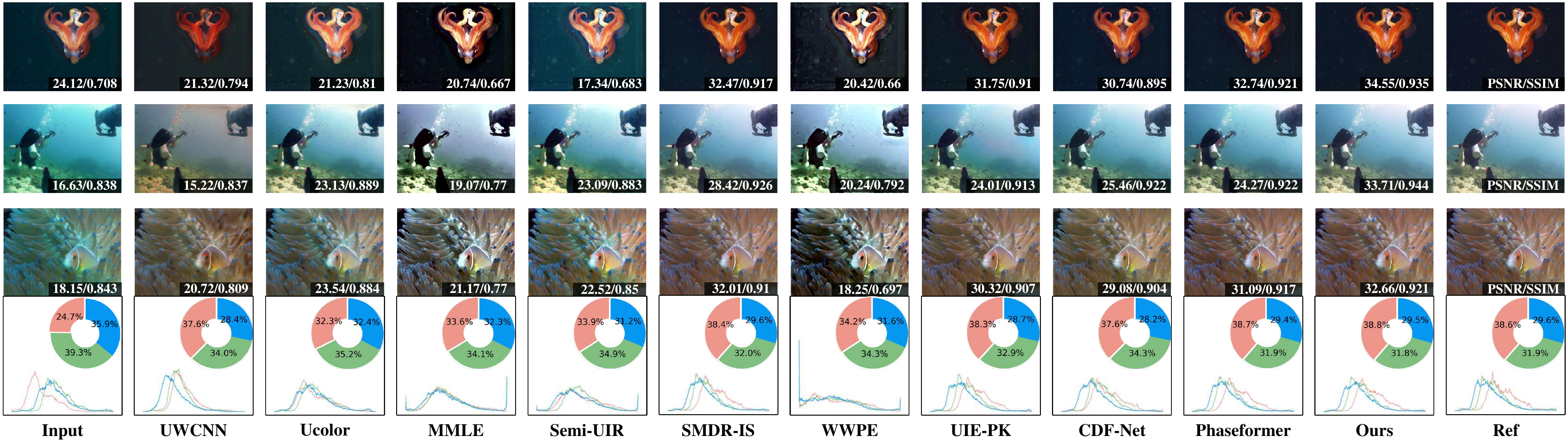}
		\caption{Qualitative comparison on the LSUI dataset. Please zoom in to inspect fine details. The PSNR and SSIM values are shown at the bottom-right corner of each image. The last row illustrates the RGB channel-wise color histograms of the third-row images, along with pie charts showing the pixel distribution across the RGB channels for each image.}
		\label{fig:EUVP}
	\end{figure*}
	
	\begin{figure*}[h]
		\centering
		\includegraphics[width=1\textwidth]{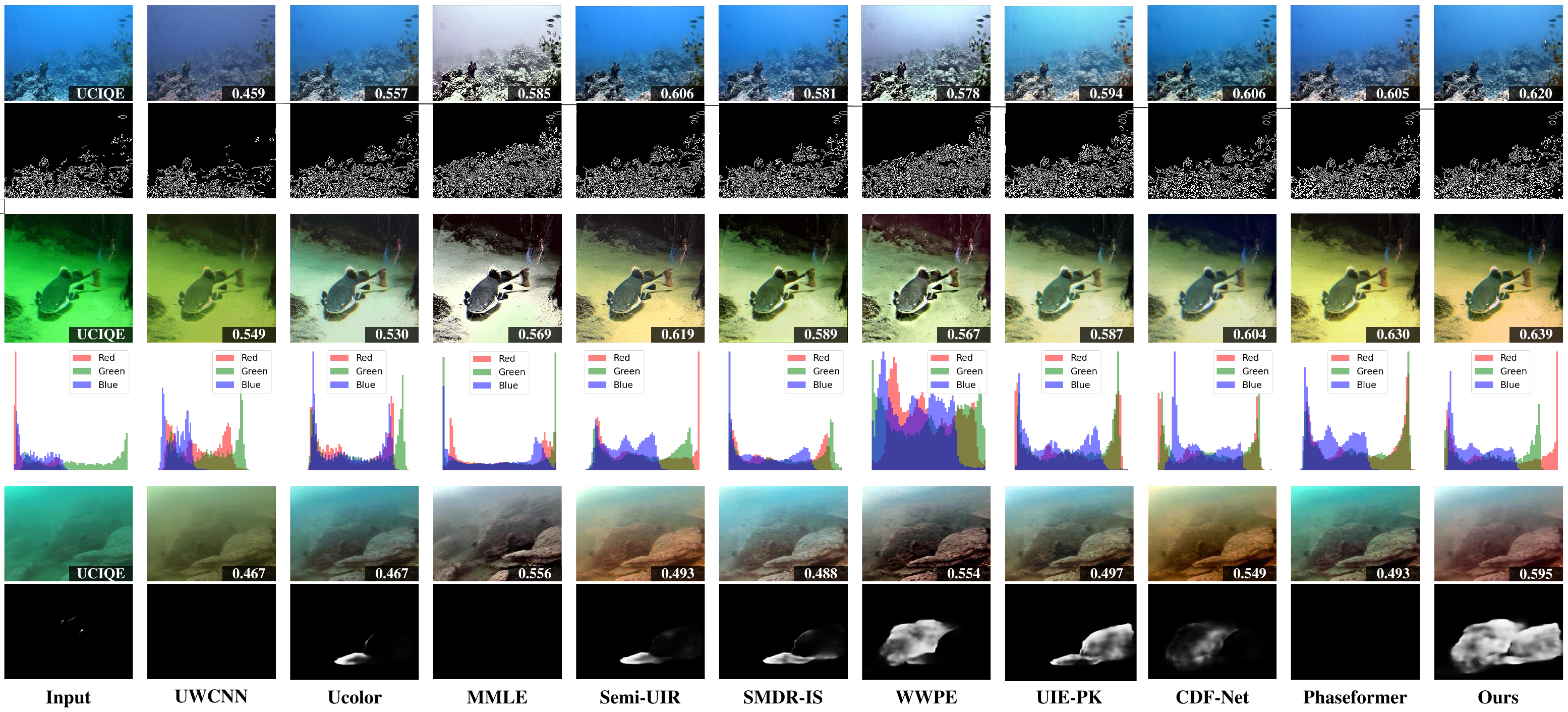}
		\caption{Qualitative comparison on the U45, UCCS, and UIEB datasets. Please zoom in to examine fine details. The UCIQE score for each image is shown in the bottom-right corner. The second row presents the RGB distribution histograms of the first-row images, the fourth row shows the saliency detection results of the second-row images, and the sixth row displays the edge detection results of the fifth-row images.}
		\label{fig:Non_Ref}
	\end{figure*}

	\subsubsection{\textbf{Qualitative Comparison}}
	Fig. \ref{fig:UIEB} and Fig. \ref{fig:EUVP} illustrate enhancement results on two paired datasets. Overall, images enhanced by UWCNN appear overly dark, while traditional methods such as MMLE and WWPE exhibit unrealistic color restoration. Additionally, the improvement provided by Ucolor is notably insufficient, especially on the UIEB dataset. Semi-UIR performs reasonably well on UIEB but exhibits poorer results on EUVP. As seen from the error map visualization in the fourth row of Fig. \ref{fig:UIEB}, the outputs from CDF-Net still exhibit significant differences compared to the reference images, indicated by regions of high error values. UIE-PK similarly performs suboptimally. In contrast, SMDR-IS, Phaseformer, and our method generate results that are closer to the reference images. Moreover, as illustrated in the fourth row of Fig. \ref{fig:EUVP}, our method produces results that most closely resemble the reference images in terms of both color histogram distributions and RGB pixel distribution. Phaseformer, UIE-PK, and CDF-Net follow sequentially in performance.
	
	Fig. \ref{fig:Non_Ref} demonstrates visual enhancement results of our method on three unpaired datasets. The first row shows results on the UIEB60 dataset. We observe that UWCNN exhibits similar characteristics to its performance on paired datasets, producing images that appear turbid. The color restorations of MMLE and WWPE are unnatural, while UIE-PK overly brightens the images. CDF-Net and Ucolor offer insufficient enhancement, resulting in persistent greenish color casts in the background. From a perceptual perspective, our method, Phaseformer, and SMDR-IS yield superior visual quality. Furthermore, considering the UCIQE scores and edge detection outcomes, the images enhanced by our method not only exhibit better visual appeal but also facilitate subsequent downstream tasks more effectively. The third row presents results from the U45 dataset. By analyzing RGB channel histograms, we specifically examine color restoration performance under extreme green-dominant scenarios. The enhanced images from our approach clearly display more natural colors, effectively compensating for the attenuation of red wavelengths underwater. The fifth row presents examples from the UCCS dataset. Saliency detection is applied to these enhanced images, as illustrated in the sixth row, where only images enhanced by our method successfully highlight both foreground reefs. Visually, our enhanced results closely resemble terrestrial images, accurately restoring the authentic colors of the reefs.
	
	Through comprehensive experiments and qualitative analyses across multiple datasets, it is evident that our enhanced underwater images not only appear closer to terrestrial scenes in terms of visual perception but also, as shown by statistical visualization and downstream task evaluations, demonstrate superior performance and higher robustness compared to recent SOTA methods.
	
	\subsection{Ablation Study}
	\subsubsection{\textbf{Ablation Study on Essential Components}}
	In order to verify the significance of the key components proposed in our SSD-Net, which are PFDB, BFCB, and AST. We conduct a series of ablation studies by systematically removing one or more components and comparing their performance, as presented in Table \ref{tab:Components}. In this table, a checkmark indicates the inclusion of the corresponding component. Specifically, it is noteworthy that in M1, M2, and M4, we replace the AST module with a conventional Transformer structure; in M4, we substitute the PFDB module with a dual-branch convolutional network of equivalent computational complexity for feature extraction; and similarly, in M1 and M3, we employ dual-branch convolutional networks with comparable computational loads to replace the BFCB module for feature refinement. As demonstrated by the results in Table \ref{tab:Components}, each proposed component significantly contributes to the overall enhancement of the model's performance.
	\subsubsection{\textbf{Ablation Study on Network Structure}}
	To validate the effectiveness of the single-scale feature extraction network structure, we modify SSD-Net into several representative multi-scale feature extraction network architectures commonly used in recent literature, as shown in Table \ref{tab:Structure}. Specifically, SSN denotes Single Scale Network, PMR denotes Progressive Multi-Resolution, and PMB denotes Parallel Multi-Branch. For a fair comparison, the primary module types of the original SSD-Net are retained during this structural transformation.
	\begin{table}[htbp]
		\renewcommand{\arraystretch}{1.1}
		\centering
		\caption{ablation studies of the proposed modules on the UIEB and EUVP datasets. A "\checkmark" indicates that the corresponding module is included in the respective model, and the \textbf{bolded} values indicate the best metric scores.}
		\resizebox{\linewidth}{!}{
			\begin{tabular}{cccccccc}
				\hline
				\multirow{2}[2]{*}{Model} & \multirow{2}[2]{*}{PFDB} & \multirow{2}[2]{*}{BFCB} & \multirow{2}[2]{*}{AST} & \multicolumn{2}{c}{UIEB} & \multicolumn{2}{c}{EUVP} \bigstrut[t]\\
				&       &       &       & SSIM↑ & PSNR↑ & SSIM↑ & PSNR↑ \bigstrut[b]\\
				\hline
				\hline
				M1    & \checkmark     &       &       & 0.862 & 22.28 & 0.858 & 25.78 \bigstrut[t]\\
				M2    & \checkmark     & \checkmark     &       & 0.915 & 24.39 & 0.911 & 26.46 \\
				M3    & \checkmark     &       & \checkmark     & 0.891 & 23.46 & 0.899 & 27.74 \\
				M4    &       & \checkmark     &       & 0.843 & 21.68 & 0.832 & 24.87 \\
				SSD-Net & \checkmark     & \checkmark     & \checkmark     & \textbf{0.924} & \textbf{24.90} & \textbf{0.917} & \textbf{29.84} \bigstrut[b]\\
				\hline
			\end{tabular}
		}
		\label{tab:Components}%
	\end{table}

	\begin{table}[h]
		\renewcommand{\arraystretch}{1.1}
		\centering
		\caption{able xxx compares the quantitative performance of various network architectures on the UIEB and EUVP datasets. Here, "3S" denotes Single-Scale Structure, and "MSS" denotes a Multi-Scale Structure, and the \textbf{bolded} values indicate the best metric scores.}
		\resizebox{0.8\linewidth}{!}{
		\begin{tabular}{cccccc}
			\hline\noalign{\vskip 2pt}
			\multicolumn{2}{c|}{\multirow{2}[1]{*}{Model}} & \multicolumn{1}{c|}{{3S}} & \multicolumn{3}{c}{MSS} \\
			\multicolumn{2}{c|}{} & \multicolumn{1}{c|}{\textbf{{SSN}}} & U-Net & PMR   & PMB \bigstrut[b]\\\noalign{\vskip 2pt}
			\hline
			\hline
			\multirow{2}[2]{*}{UIEB} & SSIM↑ & {0.924} & 0.919 & 0.923 & \textbf{0.926} \bigstrut[t]\\
			& PSNR↑ & \textbf{{24.90}} & 24.21 & \textbf{24.90}  & 24.87 \bigstrut[b]\\
			\hline
			\multirow{2}[2]{*}{EUVP} & SSIM↑ &{0.917} & 0.903 & \textbf{0.918} & 0.916 \bigstrut[t]\\
			& PSNR↑ & {29.84} & {27.87} & 29.67 & \textbf{29.88} \bigstrut[b]\\
			\hline
			\multicolumn{2}{c}{Params} & {1.026M} & 1.538M & 1.894M & 2.608M \bigstrut\\
			\hline
		\end{tabular}%
		}
		\label{tab:Structure}%
	\end{table}

	As indicated in Table \ref{tab:Structure}, the most prominent change from single-scale to multi-scale structures is the substantial increase in parameter count, ranging from a minimum increase of 49.9\% for the U-Net structure to a maximum of 160.8\% for the PMB structure. Visual inspection of Fig. \ref{fig:architecture} further demonstrates that multi-scale structures (MSS) extensively replicate similar modules across spatial resolutions or parallel branches. In contrast, the single-scale network possesses only a single streamlined backbone without redundant cross-scale downsampling/upsampling operations or lateral branch duplication, thus achieving significant parameter efficiency.
	
	Regarding performance, the single-scale architecture significantly outperforms the U-Net baseline and maintains performance comparable to PMR and PMB. This advantageous balance between performance and parameter count is attributed to our efficiently designed feature extraction module PFDB and feature fusion module BFCB. Therefore, considering both performance and parameter efficiency, the single-scale network structure represents the optimal choice.
	\subsubsection{\textbf{Ablation Study on Number of Computational Blocks}}
	Table \ref{tab:Bloks} shows ablation studies regarding the number N of cascaded PFDB and BFCB modules, and the numbe M of AST modules within each PFDB. These experiments are conducted on the EUVP dataset. By observing the changes in network parameters and PSNR values, we investigate how network performance and model complexity evolve as module numbers increase. It is evident that at smaller values of N and M, incrementing either parameter significantly improves PSNR. However, as indicated by the experimental results, the performance begins to saturate once both N and M reach 4. Further increasing the module numbers provides marginal improvements while causing a linear rise in parameter count, computation, and inference time. Additionally, deeper stacking could introduce risks of gradient vanishing, training instability, and overfitting. Therefore, considering the balance between performance and complexity, we set both N and M to 4 in our final model configuration.
	
	\begin{table}[t]
		\renewcommand{\arraystretch}{1.3}
		\centering
		\caption{Ablation results on the number N of cascaded PFDB and BFCB modules, and the number M of AST modules within each PFDB. These experiments are conducted on the EUVP dataset, and the table reports the PSNR values and corresponding model parameter counts for each configuration, the \textbf{bolded} values indicate the best metric scores, and the \textcolor{blue}{blue} entry denotes the final selected model configuration.}
		\resizebox{\linewidth}{!}{
			\begin{tabular}{c|ccccc}
				\hline
				M\textbackslash{}N & 2     & 3     & \textcolor{blue}{4}     & 5     & 6 \bigstrut\\
				\hline
				2     & 26.84/0.518M & 28.50/0.678M & 29.07/0.839M & 29.78/1.000M & 29.85/1.161M \bigstrut[t]\\
				3     & 27.13/0.549M & 28.74/0.741M & 29.36/0.932M & 29.79/1.124M & 29.85/1.316M \\
				\textcolor{blue}{4}     & {27.36/0.580M} & {28.98/0.803M} & {\textcolor{blue}{29.84/1.026M}} & {29.87/1.249M} & {29.88/1.472M} \\
				5     & 27.85/0.611M & 29.11/0.865M & 29.86/1.119M & 29.87/1.373M & 29.88/1.627M \\
				6     & 28.12/0.642M & 29.41/0.927M & 29.85/1.212M & 29.84./1.497M & \textbf{29.89}/1.782M \bigstrut[b]\\
				\hline
			\end{tabular}
		}
		\label{tab:Bloks}
	\end{table}
	\section{Conclusion}
	In this study, we revisit a long-standing assumption in underwater image enhancement—namely, that multi-scale feature fusion is essential for achieving high-quality restoration. Through both theoretical analysis and empirical validation, we demonstrate that a well-designed single-scale strategy can achieve comparable or superior performance at significantly lower complexity. To this end, we propose SSD-Net, a purely single-scale architecture that employs an asymmetric decomposition pipeline to decouple scene content from medium-induced degradation effects. The core of this architecture comprises two lightweight yet powerful modules: the Parallel Feature Decomposition Block (PFDB), which leverages an Adaptive Sparse Transformer to simultaneously separate clean and degraded features, and the Bidirectional Feature Communication Block (BFCB), which refines branch features via dynamic residual pathways while maintaining their independence. Future work will extend SSD-Net to temporally-consistent underwater video enhancement tasks and explore self-supervised training methods to alleviate the scarcity of paired training data. 
	
	\bibliographystyle{IEEEtran}
	\bibliography{reference} 
	\begin{IEEEbiography}[{\includegraphics[width=1in,height=1.25in,clip,keepaspectratio]{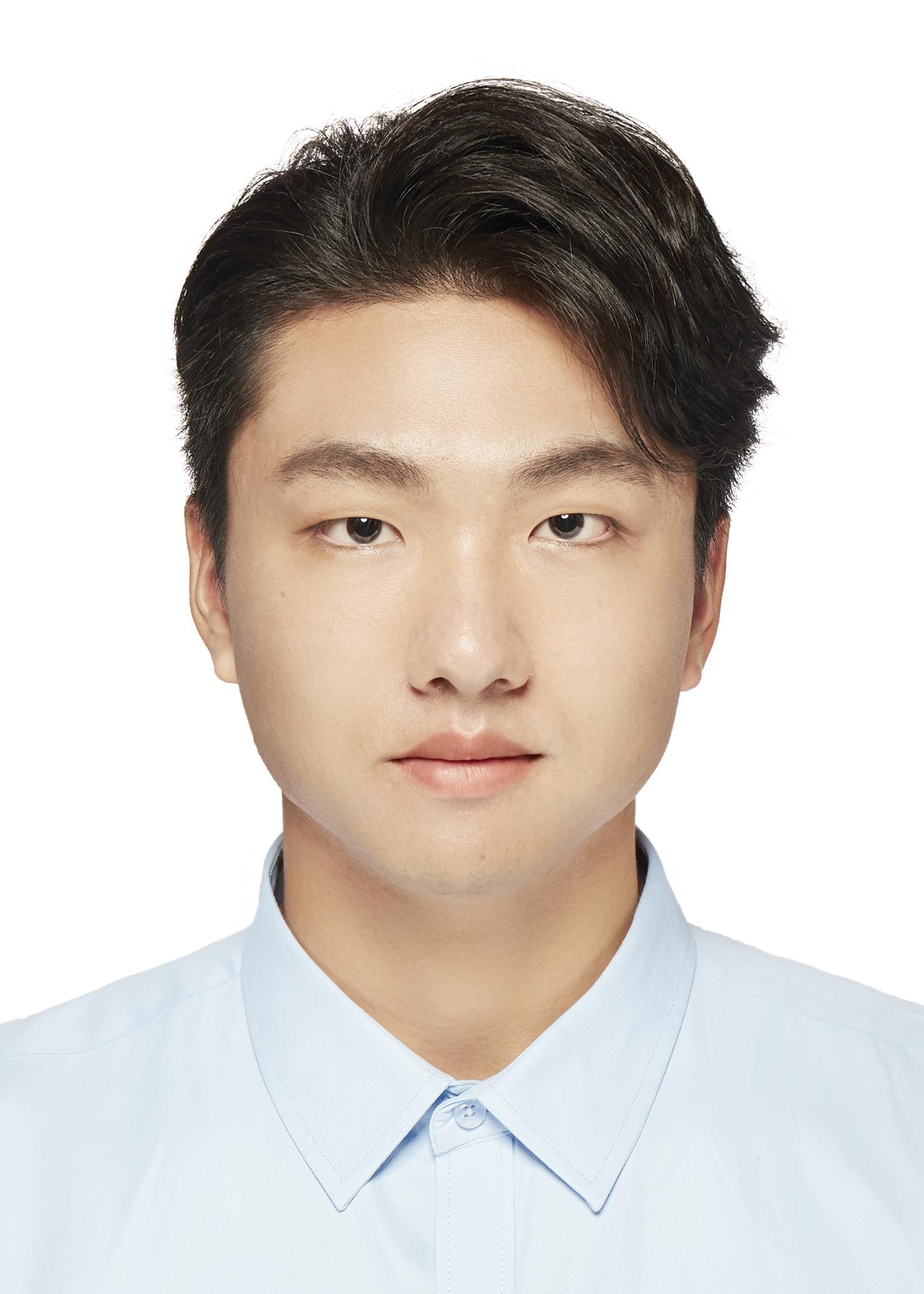}}]{Zheng Cheng}
		received the B.S. degree in Internet Engineering from Shandong University of Finance and Economics in 2019. Now he is pursuing the
		master’s degree with the College of Computer Science \& Technology, Qingdao University. His current research interests include computer vision and image processing.
	\end{IEEEbiography}

	\begin{IEEEbiography}[{\includegraphics[width=1in,height=1.25in,clip,keepaspectratio]{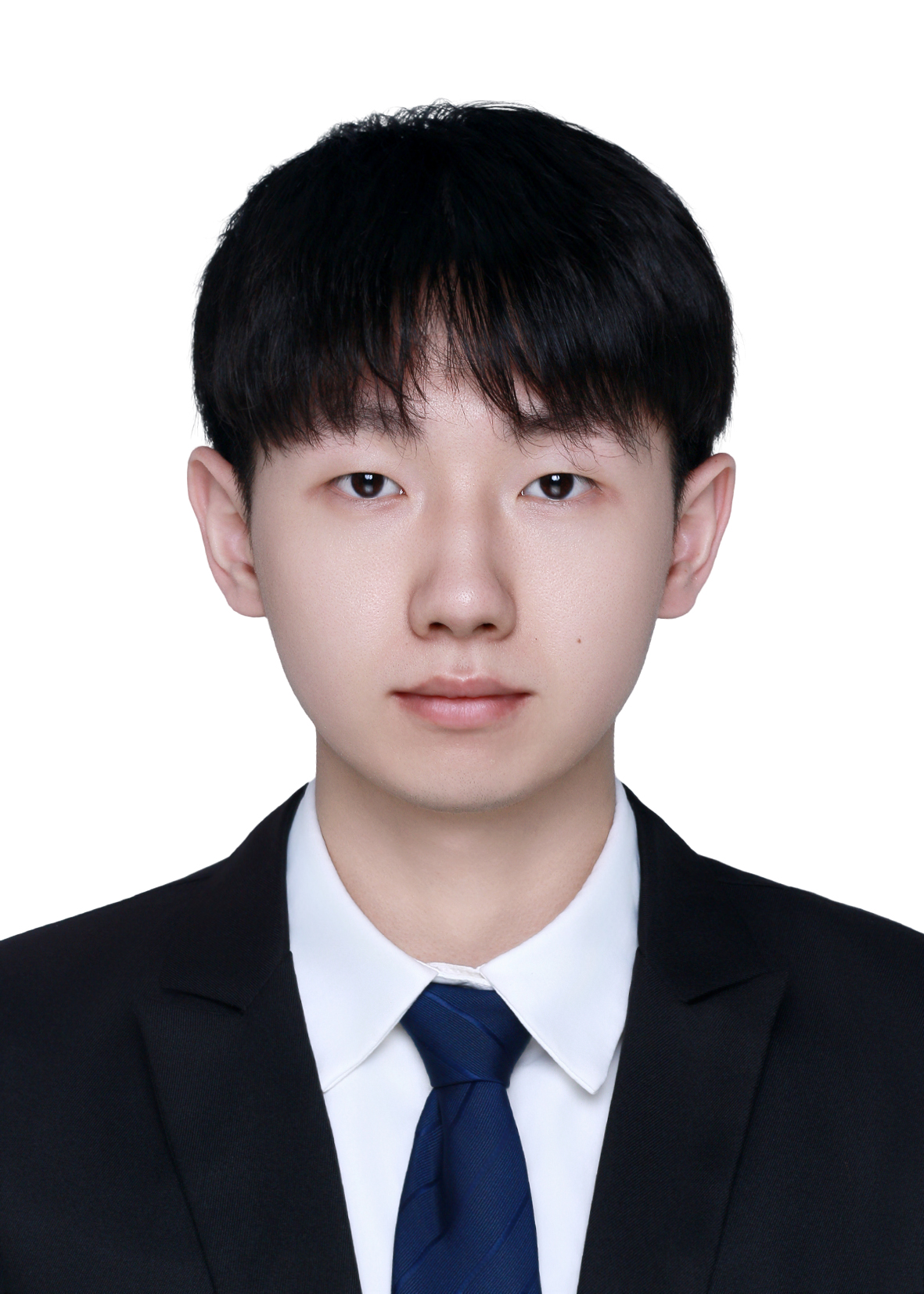}}]{Wenri Wang}
		received the B.S. degree in Data Science and Big Data Technology from Nanchang Hangkong University in 2023. Currently pursuing the M.S. degree at Fuzhou University. His research interests are in Image Enhancement , Deep Learning, and Computer Vision.
	\end{IEEEbiography}
	
	\begin{IEEEbiography}[{\includegraphics[width=1in,height=1.25in,clip,keepaspectratio]{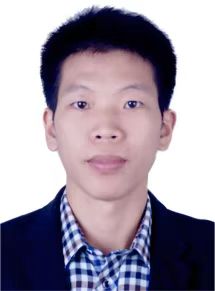}}]{Guangyong Chen}
		received the B.S. degree in mathematics from Xidian University, Xi’an, China, in 2012, and the M.S. degree in mathematics from the University of Science and Technology of China, Hefei, China, in 2014, and the Ph.D. degree in mathematics from Fuzhou University, Fuzhou, China, in 2019. His current research interests include computational intelligence, image processing, system identification, and nonlinear time-series analysis.
	\end{IEEEbiography}

	\begin{IEEEbiography}[{\includegraphics[width=1in,height=1.25in,clip,keepaspectratio]{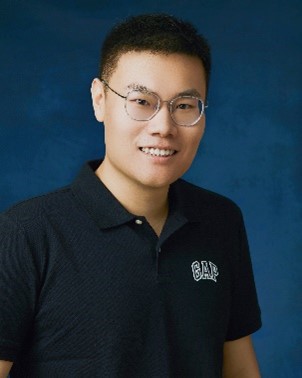}}]{Yakun Ju}
		is currently an Assistant Professor (UK Lecturer) with the School of Computing and Mathematical Sciences, University of Leicester, U.K. Previously, he served as a Postdoctoral Research Fellow at Nanyang Technological University and The Hong Kong Polytechnic University. Yakun earned his Bachelor's degree from Sichuan University in 2016 and his Ph.D. from Ocean University of China in 2022. His research interests span 3D reconstruction, medical image processing, underwater information perception, computational imaging, and low-level vision. He has authored over 60 publications in top-ranked journals and conferences, including IEEE TPAMI, IEEE TIP, IEEE TVCG, IJCV, CVPR, and NeurIPS, etc. He serves as an Associate Editor/Editorial Board for Applied Soft Computing and Neurocomputing.
	\end{IEEEbiography}

	\begin{IEEEbiography}[{\includegraphics[width=1in,height=1.25in,clip,keepaspectratio]{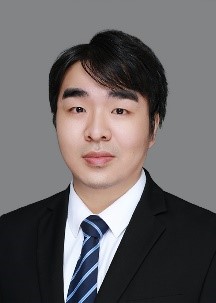}}]{Yihua Cheng}
		received the B.S. degree in computer science from Beijing University of Posts and Telecommunications in 2017, and Ph.D. degree in computer science from Beihang University in 2022. He is now a Senior Reserach Fellow with University of Birmingham, UK. His research interests include computer vision, human-computer interaction, and intelligent vehicles.
	\end{IEEEbiography}

	\begin{IEEEbiography}[{\includegraphics[width=1in,height=1.25in,clip,keepaspectratio]{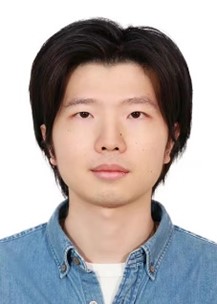}}]{Zhisong Liu}
		received his Ph.D. (2016-2020) from The Hong Kong Polytechnic University, Artificial Intelligence and Signal Processing Lab. Previously, he was a principal research scientist at DELL EMC. Before that, he was a research fellow (01/2021-12/2021) in Saint Francis University and a post-doc researcher in LIX, Ecole Polytechnique (01/2020-12/2020). His research interests include image and video signal processing, AI for atmospheric science, computer vision, and 3D data processing. He joined the CVPR Lab at LUT University in 2023 as an associate professor.
	\end{IEEEbiography}

	\begin{IEEEbiography}[{\includegraphics[width=1in,height=1.25in,clip,keepaspectratio]{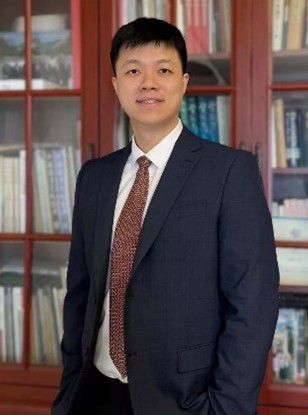}}]{Yanda Meng}
		is an assistant professor (UK Lecturer) at the University of Exeter, Computer Science Department. His research is mainly at the interface of artificial intelligence and healthcare, specialised in the research and development of artificial intelligence, biomedical image processing and analysis techniques.
	\end{IEEEbiography}

	\begin{IEEEbiography}[{\includegraphics[width=1in,height=1.25in,clip,keepaspectratio]{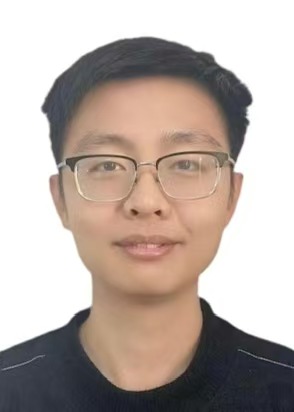}}]{Jintao Song}
		graduated with a master's degree in Computer Science and Technology from Qingdao University in 2020. Subsequently, he pursued his academic journey and obtained a doctoral degree from the same institution, Qingdao University, in 2024. Currently, his research focuses on several cutting-edge areas, including variational image processing, medical image processing, and convex optimization. 
	\end{IEEEbiography}

\end{document}